\documentclass[10pt,twocolumn,letterpaper]{article}

\usepackage{iccv}
\usepackage{times}
\usepackage{epsfig}
\usepackage{graphicx}
\usepackage{amsmath}
\usepackage{amssymb}

\usepackage[pagebackref=true,breaklinks=true,letterpaper=true,colorlinks,bookmarks=false]{hyperref}

\iccvfinalcopy 

\def\iccvPaperID{3905} 

\ificcvfinal\fi

\usepackage{amsmath}
\usepackage{amssymb}
\usepackage{booktabs}
\usepackage[noadjust]{cite}

\usepackage{float}
\usepackage{fontawesome}
\usepackage{graphicx}
\usepackage{marvosym}
\usepackage{multirow}
\usepackage{xspace}
\usepackage{wrapfig}
\usepackage{xcolor}
\usepackage{soul}
\usepackage{subfigure}
\usepackage{cleveref}
\usepackage{caption}

\definecolor{Aqua}{rgb}{0.0, 0.7, 0.7}

\newcommand{\parahead}[1]{\noindent\textbf{#1}:\ }

\newenvironment{packed_itemize}
{\begin{itemize}
    \setlength{\itemsep}{1pt}
    \setlength{\parskip}{0pt}
    \setlength{\parsep}{0pt}
}{\end{itemize}}

\newcommand{\filluptopage}[1]{%
  \clearpage
  \loop\ifnum\value{page}<#1\relax
    \null\clearpage
  \repeat
  \loop\ifnum\value{page}=#1\relax
    \null\clearpage
  \repeat
}

\makeatletter
\def\blfootnote{\xdef\@thefnmark{}\@footnotetext}
\makeatother

\begin{document}

\newcommand{\ShortName}{StrobeNet\xspace}
\title{\ShortName: Category-Level Multiview Reconstruction of Articulated Objects}

\author{Ge Zhang\\
ShanghaiTech University\\
\and
Or Litany\\
NVIDIA\\

\and
Srinath Sridhar\\
Brown University\\

\and
Leonidas Guibas \\
Stanford University\\

}

\twocolumn[{
\renewcommand\twocolumn[1][]{#1}%
\maketitle
\vspace{-0.5in}
\begin{center}
    \centering
    \includegraphics[width=\textwidth]{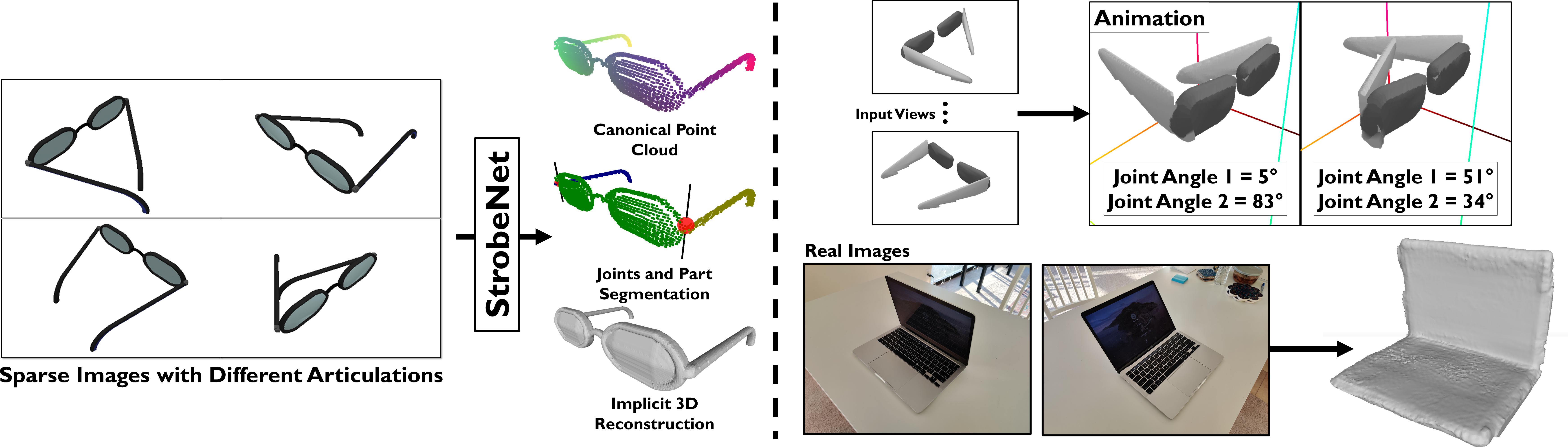}
    \vspace{-0.3in}
    \captionof{figure}
    {
    \textbf{\ShortName} reconstructs animatable 3D models of articulating objects from one or more unposed RGB images.
    Articulating objects can appear in different postures in different views (like stroboscope images).
    (Left)~Given sparse large-baseline input views, we reconstruct a feature-enriched canonical point cloud, joints and part segmentation.
    These are then used to generate a final implicit 3D reconstruction.
    (Top Right)~For previously unseen instances, we can reconstruct animatable 3D models that can be posed in any articulation.
    (Bottom Right) Our method can even reconstruct real images.
    }
    \label{fig:teaser}
\end{center}%
}]

\begin{abstract}
\vspace{-0.11in}
We present \ShortName, a method for category-level 3D reconstruction of articulating objects from one or more unposed RGB images.
Reconstructing general articulating object categories
is challenging since objects can have wide variation in shape, articulation, appearance and topology.
We address this by building on the idea of category-level \textbf{articulation canonicalization} -- mapping observations to a canonical articulation which enables correspondence-free multiview aggregation.
Our end-to-end trainable neural network estimates feature-enriched canonical 3D point clouds, articulation joints, and part segmentation from one or more unposed images of an object.
These intermediate estimates are used to generate a final implicit 3D reconstruction.
Our approach reconstructs objects even when they are observed in different articulations in images with large baselines, and animation of reconstructed shapes.
Quantitative and qualitative evaluations on different object categories show that our method is able to achieve high reconstruction accuracy, especially as more views are added.
  
  
\end{abstract}
\section{Introduction}
Reconstructing the 3D shape of everyday objects is an important problem in applications such as robotics.
To effectively operate in the real world, robots must not only understand the 3D shape of \emph{rigid} objects, but also of \emph{articulating} objects (\eg,~laptops).
There exists a large body of work on image-based 3D reconstruction of rigid objects~\cite{goesele2006multi,kar2015category,groueix2018atlasNet} as well as on articulating human bodies~\cite{saito2020pifuhd,kanazawa2018end,zhou2018monocap}.
However, reconstructing the 3D shape of previously unseen instances from common \textbf{articulating object categories} remains an open problem.

In this paper, our goal is to reconstruct \emph{an animatable} 3D model of articulating objects observed in different poses from a sparse set of one or more \emph{unposed} input images.
%
Reconstruction from sparse unposed views is an ill-posed problem that is made even harder for articulated objects since they can have wide variation in shape and appearance, and could be in different poses in different input views.
When video input with overlapping frames is available, approaches like non-rigid Structure from Motion (NR-SfM)~\cite{bregler2000recovering,akhter2008nonrigid}, or tracking~\cite{zollhofer2014real,innmann2016volumedeform} can leverage correspondences to recover 3D geometry.
However, correspondences are hard to find in images with large baselines, especially when objects articulate differently in each view.
While recent methods show promising results in such cases by learning category-level shape priors~\cite{groueix2018atlasNet,fan2017point,xu2019disn,OccNet}, they assume that objects are rigid.
%

\ShortName (named after stroboscope images) is, to our knowledge, the first learning-based method that reconstructs articulated objects as a continuous, animatable implicit function model~\cite{chibane20ifnet,OccNet,park2019deepsdf} from sparse input views (see \Cref{fig:teaser}).
Our approach is based on two key ideas.
(1)~We learn category-level shape priors for reconstructing not only rigid objects but also common articulating objects.
To enable this, we use SAPIEN~\cite{xiang2020sapien} and Shape2Motion~\cite{wang2019shape2motion}, datasets that contain common articulating object categories, and generated a new image dataset with over 120k different views of different articulating objects.
This large dataset could allow us to estimate shape by training a convolutional neural network (CNN) as a category-level shape prior.
However, the space of already large object shape and appearance variation becomes even larger with articulations making learning difficult.
%
(2)~To overcome this, we use \emph{canonicalization}, a tool for category-level 3D object understanding~\cite{Wang_2019_CVPR,xnocs_sridhar2019,novotny2019c3dpo,li2019articulated-pose} that maps object observations to a container space that is canonicalized for properties such as position, orientation and size~\cite{Wang_2019_CVPR}.
Articulation canonicalization~\cite{li2019articulated-pose} allows correspondence-free multiview shape aggregation \textbf{even when objects are observed in different articulations}.

Our network takes one or more unposed input views and estimates an intermediate articulation-canonicalized point cloud enriched with learned features, location and orientation of joints, and part segmentation of the object.
We use the aggregated 3D point cloud to drive a neural implicit function network~\cite{chibane20ifnet}, and output a smooth and continuous 3D shape.
Our network implicitly models articulation resulting in improved reconstruction quality.
Finally, our method can generate \emph{animatable} 3D models at arbitrary articulations even without using any category-specific shape templates.


%
We evaluate our approach on a newly-created dataset of 3 common articulating object categories, with full ground truth labels.
Our method compares favorably to rigid multiview aggregation methods on single view reconstruction, and significantly outperforms baselines in the articulated multiview aggregation setting.
We additionally observe that joint and part segmentation estimation improves reconstruction quality.
\ShortName provides articulation manipulation functionality which can be used to animate articulated objects.
To sum up our contributions:
\vspace{-0.1in}
\begin{packed_itemize}
    \item We introduce the problem of 3D implicit shape reconstruction of articulating objects from sparse unposed RGB views, and the first method to address this problem.
    \item A neural network architecture that explicitly models articulation to lift unposed RGB views of articulating objects to a 3D canonical point cloud and subsequently an implicit 3D shape.
    \item A method that does not require category-level templates, is fully correspondence-free, and can generate an \emph{animatable} model of articulating objects.
\end{packed_itemize}

%
\section{Related Work}
Our goal is to operate on common articulating object categories and to generalize to previously unseen instances.
While we are the first to explore this particular setting, here we discuss closely related works which we split into reconstruction of rigid, non-rigid, and human shapes from sparse images.  Since \ShortName uses multiple types of 3D shape representations, we also discuss recent advances in this field in the context of learning. 

\subsection{Multi-view reconstruction}
\parahead{Rigid Objects}
Reconstruction from images of rigid objects and scenes is a long-standing problem of interest with many classical approaches like voxel coloring~\cite{seitz1999photorealistic}, space carving~\cite{kutulakos2000theory}, or multiview stereo~\cite{goesele2006multi,furukawa2009accurate}.
These approaches assume that objects of interest are well segmented, camera poses are known, camera baselines are small, and do not make use of any category-level priors.
More recent learning-based methods overcome some of these limitations by learning category-specific priors~\cite{kar2015category,fan2017point,xu2019disn,park2019deepsdf,girdhar16b,marrnet,hsp,kanazawa2018learning,pontes2017image2mesh,deepMarching,wang2018pixel2mesh,lin2018learning,groueix2018papier,OccNet,gao2020learning} to reconstruct models from a single image.
When 3D supervision is not available, some methods move away from category-level priors instead using foreground masks~\cite{gwak2017weakly,wiles2017silnet}.
When multiple views are available at inference time, many methods explore how to exploit multiview constraints~\cite{ji2017surfacenet,kar2017learning,huang2018deepmvs,sitzmann2019scene}.
Other approaches for combining multiview information include using recurrent neural nets~\cite{ChoyXGCS16,tulsiani2020objectcentric,tulsiani2017multi}, or using feature-level aggregation~\cite{xnocs_sridhar2019,pix2surf_2020,Sitzmann_2019_CVPR}, or differentiable rendering~\cite{insafutdinov2018unsupervised,chen2019learning}.
Different from these methods, we focus on articulated objects observed from multiple views.

\parahead{Non-Rigid Objects}
Work on non-rigid object reconstruction can be divided into two categories: (1)~non-rigid structure from motion (NR-SfM), and (2)~non-rigid tracking approaches.
NR-SfM methods typically focus on reconstructing sparse points~\cite{akhter2010trajectory,dai2014simple,agudo2018image}, or use basis shapes, priors, or low-rank approximations to overcome the under constrained nature of the problem~\cite{bregler2000recovering,akhter2008nonrigid,zhu2014complex,torresani2008nonrigid,bartoli2008coarse}. Learning morphable models from images has recently received increasing attention~\cite{kanazawa2018learning,tulsiani2020implicit,li2020self} together with learning 4D reconstruction~\cite{niemeyer2019occupancy}.
Tracking approaches typically operate on \mbox{RGB-D} videos and either assume that a template of the instance is available~~\cite{innmann2016volumedeform,golyanik2017accurate,guo2015robust,hernandez2007non,li2009robust}, or that neighboring frames have short baselines~\cite{newcombe2015dynamicfusion,slavcheva2017killingfusion,zollhofer2014real,li2020online}.
Our focus is on general articulating object categories that are observed in RGB images with large camera baselines, and have varying topologies and diverse shapes that make it difficult for the above methods.


\parahead{Humans}
Human bodies, hands, and faces are special cases of non-rigid articulating objects and have received special attention.
Like many non-rigid reconstruction methods, methods for human shape rely on template models, priors, and the fixed topology of human shapes.
Extensive work has been done on using multiple views for shape reconstruction including motion~\cite{de2008performance,starck2007surface,ballan2012motion}.
These methods assume studio settings with numerous \emph{synchronized} cameras, a template model, and fixed kinematic skeletons.
Depth sensors can alleviate the need to have multiple synchronized cameras but many methods still use templates~\cite{baak2013data,ganapathi2010real,tkach2016sphere}, with help from parametric body models~\cite{loper2015smpl,anguelov2005scape}.
More recently, deep learning methods has successfully shown shape reconstruction from videos~\cite{alldieck2018detailed,alldieck2018video} but they assume that the person does not move or change their articulation during capture.
For single views, there exist methods for high-quality reconstruction~\cite{saito2020pifuhd,saito2019pifu,kanazawa2018end,zhou2018monocap,habermann2020deepcap}, but fusion of multiple views of different poses remains an open problem.
Our goal of reconstructing animatable shape is close to that of ARCH~\cite{huang2020arch}, although they rely heavily on body priors.

\parahead{Shape Representations}
There exist many choices for representing 3D shape from point clouds~\cite{fan2017point}, depth images~\cite{zhang2018learning,xnocs_sridhar2019}, voxel grids~\cite{seitz1999photorealistic}, explicit meshes~\cite{de2008performance,ballan2012motion}, or implicit surfaces~\cite{park2019deepsdf,OccNet,chibane20ifnet}.
Methods based on implicit functions have shown promise in faithfully representing complex 3D shapes and are widely used in learning-based reconstruction~\cite{park2019deepsdf,OccNet}.
Recently, ~\cite{Hao_2020_CVPR} proposed a dual representation that can be realized as either coarse manipulable pointclouds or smooth implicit surface. Similarly, in our work, we adopt two representations: a canonical depth map to represent partial object shapes from multiple views~\cite{xnocs_sridhar2019} that can be easily animated, and implicit functions that faithfully reconstruct articulated shape~\cite{chibane20ifnet}.
\section{Background}
\label{sec:back}
\parahead{Canonicalization}
Canonicalization refers to the task of mapping object observations (pixels or points) to a canonical container that normalizes object properties like position, orientation and size.
It has been used recently as a tool for solving category-level 3D object understanding tasks such as 6~DoF pose estimation~\cite{Wang_2019_CVPR,li2019articulated-pose}, automatic labeling~\cite{zakharov2020autolabeling}, and 3D reconstruction~\cite{xnocs_sridhar2019,novotny2019c3dpo,novotny2020canonical}.
Canonicalization can be learned using full supervision~\cite{Wang_2019_CVPR} or without using any supervision~\cite{novotny2019c3dpo} -- we follow the supervised approach in this work.
Specifically, we adopt Normalized Object Coordinate Space (NOCS)~\cite{Wang_2019_CVPR} as our representation.

\begin{figure}[ht!]
    \centering
    \includegraphics[width=\columnwidth]{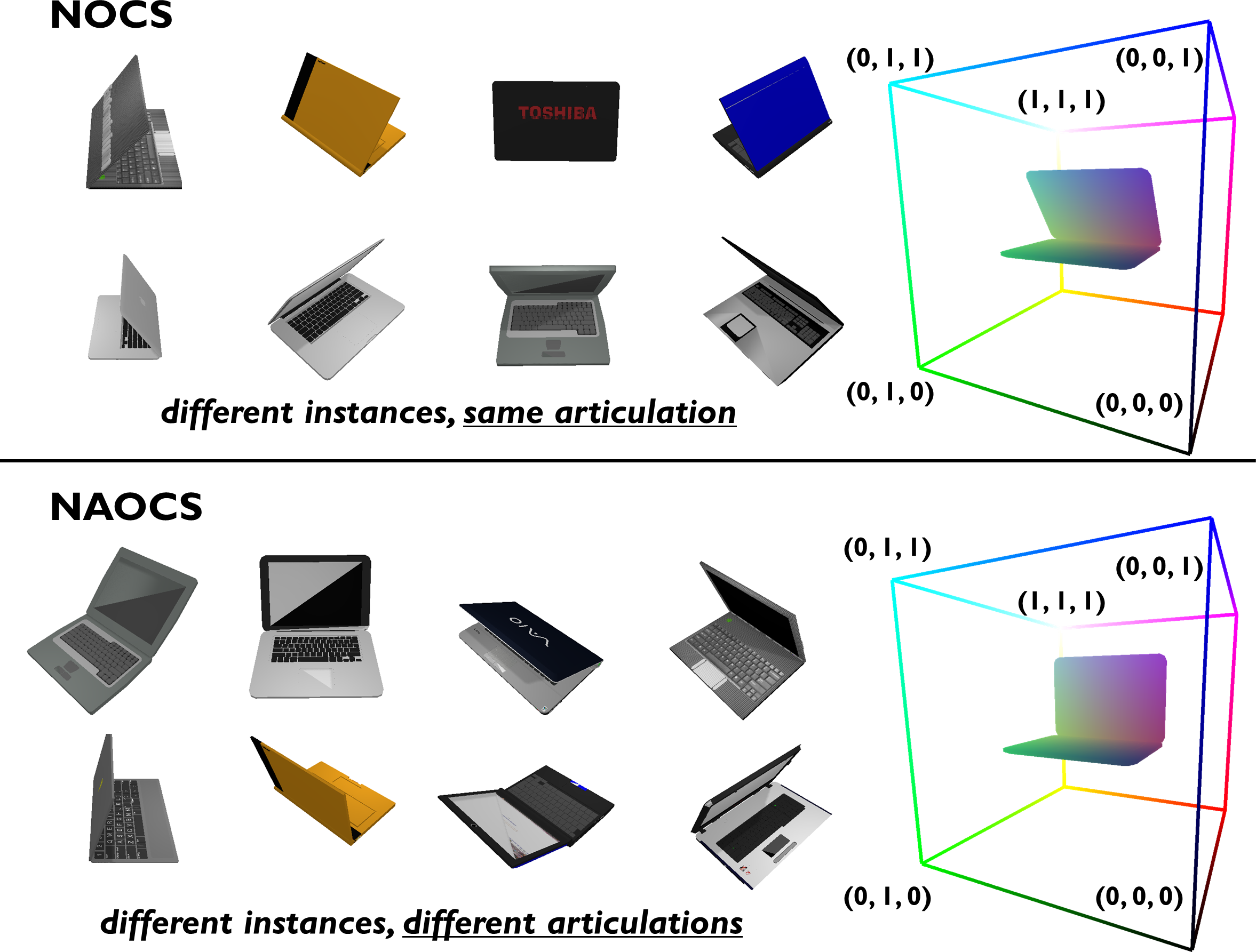}
    \caption{
    (Top)~NOCS~\cite{Wang_2019_CVPR} is a canonical container that maps different object instances to a unit cube with consistent position, orientation, and size.
    However, the articulation is left unchanged (laptop lid is articulated).
    (Bottom)~NAOCS~\cite{li2019articulated-pose} is a container that additionally canonicalizes for articulation (laptop lid is set to a pre-determined pose).
    This allows the aggregation of shape even when they are in different articulations.
    The RGB colors denote the (x,y,z) position of a point within the container.
    }
    \label{fig:canon}
\end{figure}

\parahead{NOCS and NAOCS}
NOCS is a standard 3D container in which all instances of a certain category are consistently positioned, oriented and scaled to fit within a unit cube.
Figure~\ref{fig:canon} (top) illustrates NOCS and how it maps different laptop instances with the same articulation to a container that canonicalizes for position, orientation and size.
NAOCS~\cite{li2019articulated-pose} is an extension of NOCS to canonicalize for articulation in addition to position, orientation and scale.
Figure~\ref{fig:canon} (bottom) illustrates NAOCS and how it maps different laptop instances with different articulations to a pre-determined articulation.
Doing so allows us to aggregate shape even if individual observations are in different poses.
\vspace{-0.25in}
%

\parahead{NOCS/NAOCS Map}
A NOCS/NAOCS map is a projection of a 3D shape in the container onto a 2D image where each projected pixel carries the (x,y,z) position of the corresponding 3D point (denoted by RGB colors).
These maps encode the partial shape of an object and are equivalent to a dense point cloud.
NOCS/NAOCS maps provide a way to estimate 3D shape as 2D images within neural networks without dealing with 3D network designs.
Additionally, if joints and part segments are available, NOCS maps can be \emph{reposed} to create NAOCS maps.

We treat the 3D shape estimation task as finding a function $F\colon \mathcal{R}^{H \times W \times 3} \to (\mathcal{R}^{H \times W \times 3}, \{0,1\}^{H \times W})$ defined on RGB images that predicts a per-pixel 3D coordinate as well as a binary foreground-background mask.
Here $W$ and $H$ denote the width and height of the image.
Like previous work, we realize $F$ via a 2D CNN~\cite{Wang_2019_CVPR,xnocs_sridhar2019,li2009robust}, and train it using ground truth supervision.
We also predict the position and orientation of joints and part segments which allows us to repose NOCS maps to NAOCS maps.
When multiple views are available, we can aggregate object shape into the same NAOCS container irrespecive of the original articulation.



\begin{figure*}[!ht]
\begin{center}
\includegraphics[width=\linewidth]{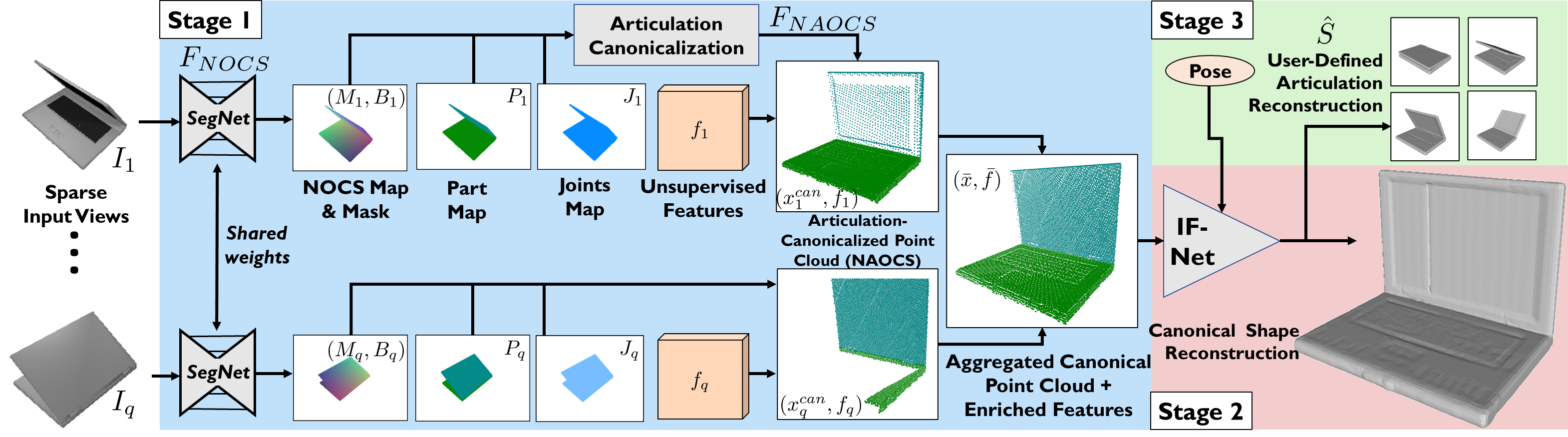}
\end{center}
\caption{
Our method takes as input a set of sparse images \(I_q\) and aims to reconstruct an animatable 3D shape \(\hat{S}\). In stage 1, we use a SegNet-like architecture to predict: (1)~the point cloud $x_q$ in NOCS as a NOCS map and mask $(M_q, B_q)$, (2)~the part segmentation $P_q$, and (3)~joint location and orientations in NOCS $J_q$. The network also outputs a high-dimensional unsupervised feature $f_q$. With the predicted part segmentation and joints, we can transform the point cloud into canonical articulation in NAOCS where inputs from arbitrary pose are aggregated ($\bar{x}$) along with enriched features $\bar{f}$. $(\bar{x}, \bar{f})$ are then fed into stage 2, an IF-Net, to output an reconstruction at any resolution, or articulation (stage 3).}
\label{fig:pipeline}
\end{figure*}

\parahead{Alternative Formulations}
Different from us, \cite{li2019articulated-pose} used per-part NOCS containers (referred to as NPCS) for articulated object pose estimation.
While this is important in their method, there are several reasons why we choose a different formulation.
First, their approach solves a different problem using 3D point clouds from depth images which are less noisy than lifting from 2D images.
Thus they are able to use the shape of small object parts -- this can be challenging for shapes lifted from 2D.
Second, our approach is scalable to any articulating object category since we do not require a per-part learning allowing future extension of our method.



\section{Method}
\label{sec:method}
Given a sparse set of RGB views of an object $\{I_q \in \mathcal{R}^{H \times W \times 3}\}_{q=1}^Q$, possibly at different articulations in each view, our goal is to reconstruct the 3D shape $\hat S$ of that object at arbitrary articulations.
Achieving this requires not just estimating shape but also joints positions and orientations, and part segmentation for reposing objects.
We also need to handle different articulations in input views -- we use a category-level canonical 3D space to which the input views are lifted while stripping away the view-specific scale, pose, and articulation information.
Our approach, depicted in Figure~\ref{fig:pipeline}, has 3 stages: (1)~lifting 2D views to a canonical 3D space, (2)~surface reconstruction, and (3)~reposing.
We describe each of these stages in detail below.

\subsection{Stage 1: Lifting to a Canonical 3D Space}
\label{section:pnnocs}
To lift the RGB input images to 3D, we realize the function $F_{NOCS}(I_q) = \{M_q, B_q\}$ described in Section~\ref{sec:back} using a CNN and 
%
%
estimate NOCS maps $M_q$ with per-pixel 3D coordinates along with a binary object mask $B_q$.
These grant us an equivalent representation in the form of a point cloud ${x_q} = \{M_q(w,h) | B_q(w,h)=1 \} \in \mathcal{R}^{N_q\times 3}$, where $N_q=\Sigma_{(w,h)} B_q(w,h)$ is the number of valid foreground pixels of the $q$-th view, and $(w, h)$ are the image width and height indices. 
While $\{x_q\}_{q=1}^Q$ are canonicalized for position, orientation, and size, we still cannot use them to perform multiview aggregation directly as each view presents the object in a different articulation. We therefore adopt the articulation-canonicalized maps (NAOCS)~\cite{li2019articulated-pose}.

Instead of directly estimating NAOCS we structure the prediction by explicitly modeling the notion of articulation into the network.
We do this by making the network estimate object parts and joints in addition to NOCS maps. 
We then use the parts and joints to \textbf{repose} the predicted NOCS map into a NAOCS map. 
Our structured prediction has two advantages.
First, it enables an explicit access to the object pose thus allowing to conveniently animate it.
Second, by supervising the reconstructions at all input articulations we avoid overfitting to the canonical one and strengthen the generalization capabilities of our network,
as we show in Table~\ref{tab:input_pose}.

\parahead{Joint and Part Estimation}
To facilitate articulation canonicalization and implicit modeling of articulations, we extend our map to $F_{NAOCS}(I_q)=\{M_q, B_q,P_q^1,..,P_q^{N_P},J_q^1,..,J_q^{N_J}\}$ where we further predict for each view $N_P$ binary part segmentation masks 
$P_q \in \{0,1\}^{H\times W}$ 
as well as a prediction of each joint location and its rotation angles with respect to a pre-defined rest pose $J_q \in \mathcal{R}^6$ 
%
%
for each joint $J^j$ in an object category, and foreground pixel $p$ we predict a 6D vote $v_q^j(p) = (c_q^j(p), r_q^j(p))$ along with a 1D confidence score $w_q^j(p)$.
The first 3 dimensions $c_q$ correspond to the joint center location, followed by 3 dimensions $r_q$ for joints orientation in axis angle form. 
%
%
\begin{figure}[th!]
    \centering
    \includegraphics[width=\columnwidth]{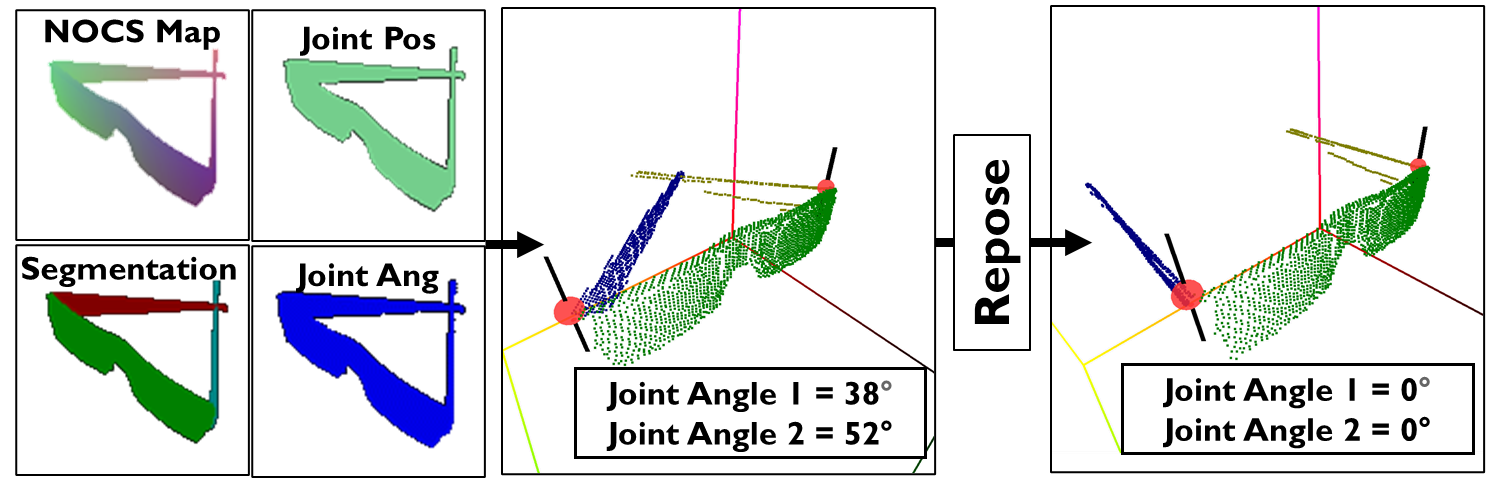}
    \vspace{-0.15in}
    \caption{\textbf{Articulation canonicalization module}. Given estimates for NOCS map $M_q$, part segmentations $P_q$, and the aggregated joint location and joint angles $J_q$ (left) resulting in a canonical, segmented and animatable 3D poincloud (center) our module ``unrotates'' the parts back to their rest pose shape $x_q^{can}$ (right).}
    \label{fig:art_canon}
    \vspace{-0.15in}
\end{figure}
We additionally output $C$-dimensional per pixel features (unsupervised) $f_q(p)$ which will be used in stage 2.
Our approach only requires the number of joints and parts to be specified \textbf{once per category} (our dataset contains this information), and does not require a template shape.
We note that \cite{li2019articulated-pose} used a different parameterization of pivots and angles, and predicted a per-point joint association. Instead we retrieve the association directly from the part segmentation and allow all points to participate in the vote aggregation.


\parahead{Articulation Canonicalization}
%
Articulation canonicalization allows reconstructions from multiple views and poses to be aggregated.
We now explain the reposing of the parts to generate the articulation-canonicalized 3D point cloud.
Specifically, to each lifted point $p \in x_q$ that belongs to part $P^i$ (and is thus associated with joint $J^j$) we apply a rotation by the inverse of the aggregated predicted rotation $\Sigma_p w_q^j(p)r_q^j(p)$ around the joint center $\Sigma_p   w_q^j(p)c_q^j(p)$.
The result is an estimated per-view articulation-canonicalized 3D point cloud ${x_q^{can}} \in \mathcal{R}^{N_q\times 3}$, enriched with features ${f_q}\in \mathcal{R}^{N_q\times C}$ (see Figure~\ref{fig:art_canon}).

\parahead{Multi-view Aggregation}
\label{section:reconstruction}
%
To make the best use of partial information from different views, we need to combine multi-view information during the reconstruction process.
This is not straightforward in general because the views are unposed.
However, our approach inherently supports it -- since all the per-view predicted point clouds $\{x^{can}_q\}$ are in canonical articulation, we can simply aggregate all of them using a set union operation $\bar{x} = \bigcup\{x^{can}_q\}_{i=1}^{Q}$.
Figure~\ref{fig:pipeline} shows how the shapes reconstructed from different views complement each other to reconstruct object geometry without using explicit correspondences.
Note that in addition to the 3D coordinates, each point in the aggregated cloud carries a high-dimensional feature, which we aggregate as well: $\bar{f} = \bigcup\{f_q\}_{i=1}^{Q}$. Aside from these features, all other estimated parameters are fully supervised as will be explained next.


\parahead{Network, Supervision and Losses}
We use a SegNet~\cite{badrinarayanan2017segnet} based encoder-decoder architecture with skip connections.
The network estimates NOCS maps, joints, and part segmentation, all of which are fully supervised using labels in our dataset (see Section~\ref{subsec:dset}).
Stage 1 of our architecture uses a loss function $\mathcal{L} = \Sigma_{i=1}^7 \lambda_i \mathcal{L}_i$ consisting of 7 terms.
The per-view NOCS map loss $\mathcal{L}_1$ minimizes the L2 distance between the predicted 3D coordinates of foreground pixels. 
The predicted foreground binary mask $B$ is supervised using a binary cross entropy loss $\mathcal{L}_2$.
For the dense 6D joints prediction, we estimate a $N_J \times H \times W \times 6$ channel.
The joint position loss $\mathcal{L}_3$ minimizes the per-pixel L2 distance between the first 3 channels, where the joint orientation loss $\mathcal{L}_4$ minimizes the L2 distance for the last 3 channels.
An additional joint consistency loss $\mathcal{L}_5$ minimizes the distance between the per-view joint estimates in a pair-wise fashion. 
This provides sufficient supervision to learn meaningful \emph{confidence maps} $w_q^j$ indirectly. 
For part segmentation, we use a standard multiclass cross entropy loss $\mathcal{L}_6$.
Finally, we also have a loss $\mathcal{L}_7$ between the estimated per-view \textit{canonical} point cloud $x^{can}_q$ and the ground truth.
All losses have a weight of 1.0, except the BCE loss which has a weight of 0.1.


\subsection{Stage 2: Reconstruction}
\label{section:reconstruct}
The output from stage 1 is an aggregated 3D point cloud enriched with features. 
However, this is insufficient for many downstream applications in animation and rendering for which we seek an implicit (occupancy) or explicit (mesh) 3D reconstruction.
Such reconstruction has the added benefit of smoothing out outliers and noise in reconstruction point clouds.
Therefore, the second stage of our method takes an input point cloud (with unsupervised features) and produces an smooth implicit reconstruction from which we can extract a mesh.
We build on top of IF-Net~\cite{chibane20ifnet} which was recently proposed for reconstruction from point clouds in the form of an occupancy function.
A compelling property of IF-Net is its hierarchical structure which, differently from other solutions in this category of neural implicits, allows reasoning about object parts.
This is especially important when dealing with articulations since while parts maintain the same geometry, the global structure changes entirely and may lead to poor generalization.

We modify IF-Net to support our problem using additional input channels.
Specifically, the input to IF-Net is the aggregated articulation-canonicalized point cloud enriched with features $(\bar{x}, \bar{f}) \in \mathcal{R}^{\Sigma N_q \times (3+C)}$. We discretize our unit-length canonical container into a $R\times R\times R$ voxel-grid, where each voxel contains a $C$-dimensional feature that is the average of all the points contained in that voxel.
Following IF-Net, a hierarchical point encoding $(F_1(\bar{x_i}),...,F_K(\bar{x_i}))$ is performed via a 3D CNN and fed to a point decoder $f(x_i): F_1(\bar{x_i}),...,F_K(\bar{x_i}) \to [0,1]$.
All reconstructions shown in the paper use the output occupancy processed to a mesh using marching cubes~\cite{lorensen1987marching}.

Now the goal of the additional features becomes clear.
Different from the setting in IF-Net where the (possibly partial) point-cloud is accurately located, our point locations may suffer from inaccuracies due to errors in either of the stages in the articulation canonical point cloud prediction.
We train both stage 1 and 2 of our method end-to-end which allows these additional features to be used to improve 3D shape and acts as an implicit denoising mechanism.

\parahead{Network, Supervision and Losses}
The second stage uses a single-view reconstruction version IF-Net with 11 3D convolutional layers with different granularity and 3 fully-connected layers(more details are in the supplemental). The occupancy is fully supervised by labels our dataset (see Section~\ref{subsec:dset}).
We use a binary cross-entropy loss between the ground truth occupancy and our estimate.


\subsection{Stage 3: Reposing and Animation}
\label{section:repose}
Using the above two stages, we can obtain both the aggregated articulation-canonicalized point clouds as well as a complete 3D reconstruction in implicit or explicit form.
However, our approach provides even more flexibility -- we can \textbf{repose and animate} the reconstructed object to an articulation of our choice.
We train stage 2 of our network to reconstruct the original ground truth articulation shape of the input views in addition to the articulation-canonicalized shape.
At inference time, we allow the user to freely choose a per-joint rotation angle as a natural reposing interface (see Stage 3 in Figure~\ref{fig:pipeline}).
When this rotation angle is input to stage 2, we can reconstruct the object in arbitrary articulations.
We also propose an alternative strategy for more consistent animation that takes the predicted joints and parts to repose the canonical mesh using a standard rigid skinning procedure (more details in the supplementary materials).
\section{Results \& Experiments}
\label{sec:results}
We evaluate \ShortName on our newly generated benchmark of articulated shapes and their multi-view images. After describing the curation of the dataset and supervision signals (\ref{subsec:dset}) we describe the baseline methods in comparison (\ref{subsec:baseline}). We then present the performance of our \ShortName under different settings (\ref{subsec:mainresults}), followed by an ablation study of key ingredients of our method (\ref{subsec:ablation}). Finally, we demonstrate the reposing of reconstructed objects in \ref{subsec:reposing}.


\subsection{Datasets}\label{subsec:dset}
We utilize two data sources of manipulable 3D piece-wise rigid objects to build our dataset: SAPIEN~\cite{xiang2020sapien} and shape2motion~\cite{wang2019shape2motion}.
Both provide a medium-sized collection of objects from different categories with controllable part articulation.
We generated a dataset of 3 object classes (common in related work~\cite{park2019deepsdf,xnocs_sridhar2019}): `eyeglasses'', ``laptop'' and ``oven'' at various combinations of 6D pose and articulation.
We note that in each category, all objects share the same number of parts and joints.
We randomly sample 20 part-pose combinations per object and render it from 8 different view angles and ranges.
We split our dataset between (48, 47, 26) training shapes of glasses, laptops, and ovens, respectively, and a  test set of unseen (5, 5, 3) shapes.
Due to the relatively fewer shapes, we further augment the 3D models using non-uniform scaling, and images gamma correction resulting in a total of (42400, 33280, 46400) samples per category. 




%

\subsection{Baseline methods}\label{subsec:baseline}
\begin{figure}
\begin{center}
\includegraphics[width=\linewidth]{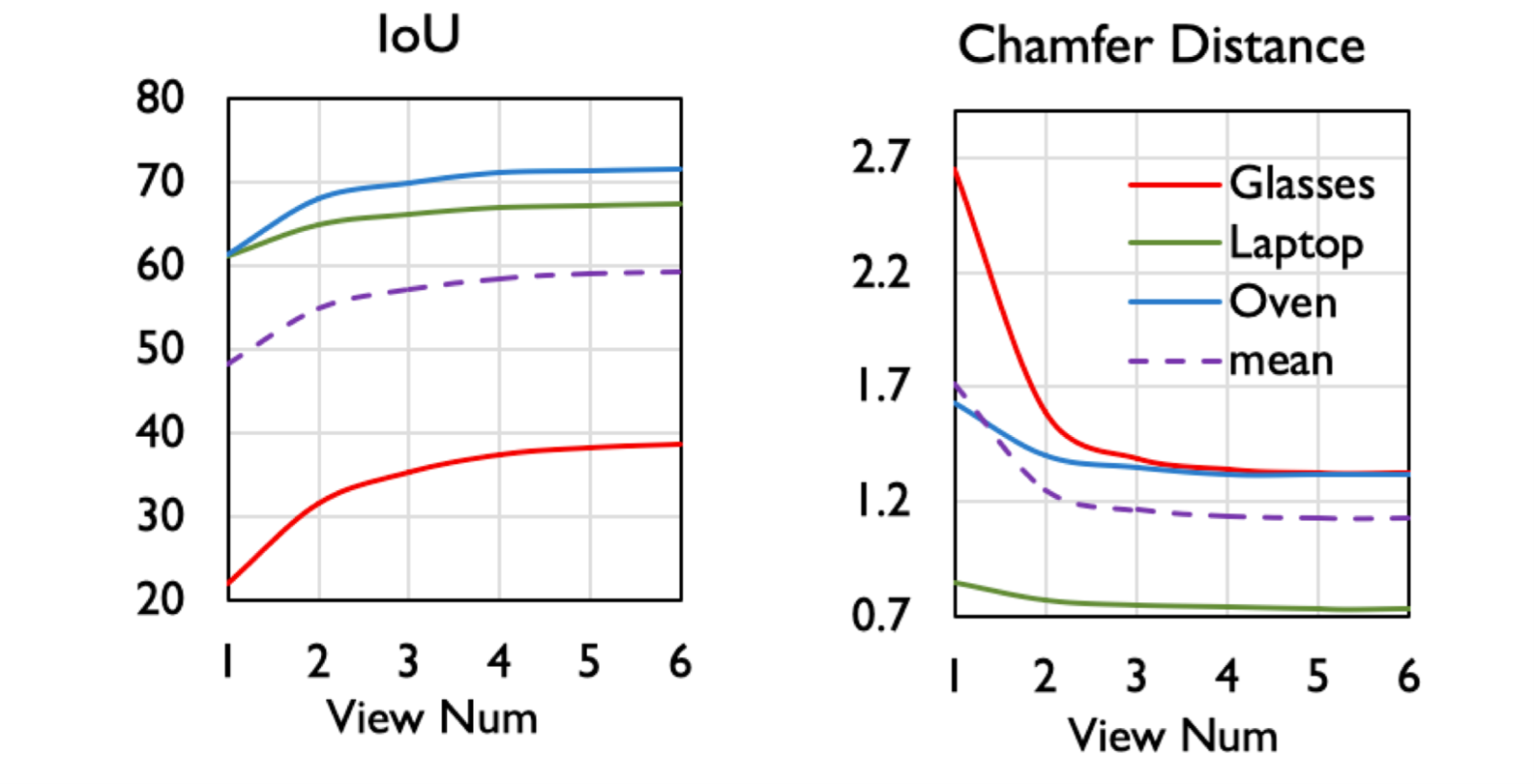}
\end{center}
   \caption{\textbf{Multiview Aggregation} \ShortName trained on 2 views could successfully infer inputs from more views, while improving reconstruction accuracy.}
\label{fig:multiview_inference_compact}
\end{figure}

To the best of our knowledge, there are no methods that are directly applicable to this new problem setting.
We therefore resorted to the closest related setting and chose OccNet~\cite{OccNet} as a representative state-of-the-art single-view reconstruction method.
To modify it for our setting, we trained two variants of OccNet on the articulated objects dataset.
\textbf{OccNet-F} is trained to reconstruct the objects in the articulation presented at the input view.
When more than one view is provided, we stack the input views and supervise using the 3D shape in the articulation of the \textit{first} view in the stack.
\textbf{OccNet-C} is instead trained to predict the 3D object in its articulation \textit{canonical} state. 
We train 3 models of each network variant on 1, 2 and 4 input views respectively and compare their reconstruction with our \ShortName. Note that while OccNet-C and OccNet-F are trained to aggregate geometry from multiple views, their architecture is inherently limited compared with \ShortName in two ways: (a)~they cannot generalize to an arbitrary number of views at inference (we demonstrate this in Figure \ref{fig:multiview_inference_compact}), and (b)~their output articulation is not animatable.
%
\begin{figure}[hb!]
\begin{center}
\includegraphics[width=1.01\linewidth]{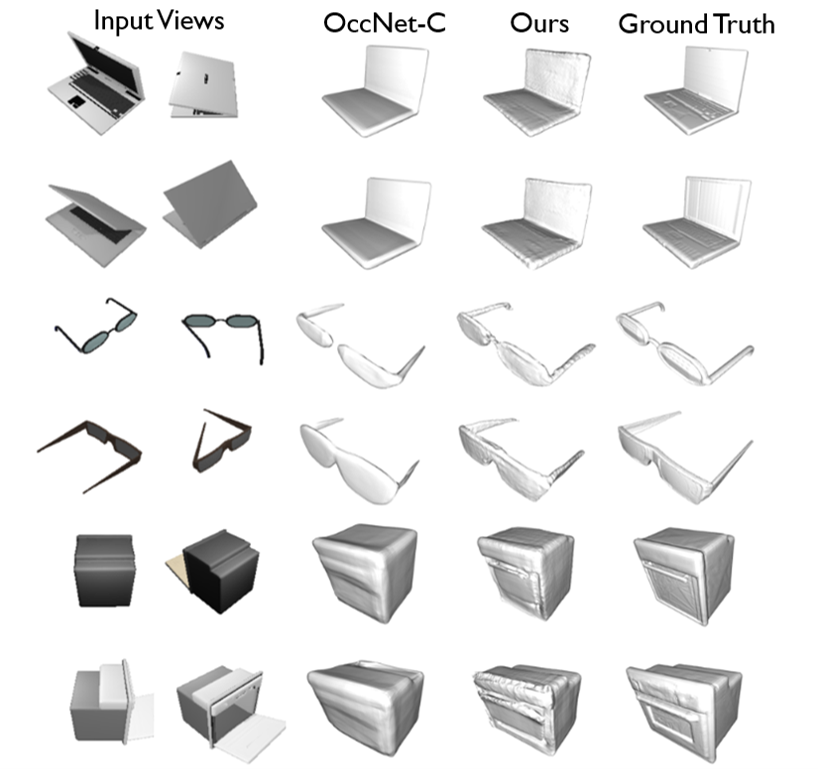}
\end{center}
   \caption{Qualitative results with 2 input views. From left to right: 2 input views, OccNet-C (baseline), StrobeNet (Ours) and ground truth. \ShortName produces better reconstructions with fine-grained details.
   }
\label{fig:visualresults}
\end{figure}

\subsection{Articulated Shape Aggregation}\label{subsec:mainresults}

\renewcommand{\arraystretch}{1.4}
\setlength{\tabcolsep}{5pt}
\begin{table*}[t!]
\centering
\centering
\footnotesize
\resizebox{0.9\linewidth}{!}{
\begin{tabular}{llcccccc|cc}
    \toprule
    &\textbf{Class} & \multicolumn{2}{c}{\textbf{eyeglasses}} &  \multicolumn{2}{c}{\textbf{laptop}} &  \multicolumn{2}{c}{\textbf{oven}}    & \multicolumn{2}{c}{\textbf{mean}} \\    
    &\textbf{Metric} & \textbf{CD~$\downarrow$} & \textbf{IoU~$\uparrow$} & \textbf{CD~$\downarrow$} & \textbf{IoU~$\uparrow$} & \textbf{CD~$\downarrow$} & \textbf{IoU~$\uparrow$} & 
    \textbf{CD~$\downarrow$} &
    \textbf{IoU~$\uparrow$}
    \\
    \midrule

    \multirow{2}{*}{\textbf{Single view}}&\textbf{OccNet-C}  &  5.38 $\pm$ 3.11 & 22.94 $\pm$ 11.53 & 1.31 $\pm$ 0.27 & 50.92$\pm$10.04 & 2.31 $\pm$ 0.59 & 50.46 $\pm$ 13.16 & 2.96 & 41.44 \\
    
    &\textbf{\ShortName (Ours)} & \textbf{3.72 $\pm$ 2.88} & \textbf{24.45$\pm$9.04} &  \textbf{0.83 $\pm$ 0.21} & \textbf{60.77 $\pm$ 10.36} & \textbf{1.42 $\pm$ 0.53} & \textbf{63.75 $\pm$ 12.04}  & \textbf{2.55} & \textbf{49.66} \\
    \hline
    
    \multirow{2}{*}{\textbf{2 views}}&\textbf{OccNet-C} & 4.47 $\pm$ 2.63 & 23.75 $\pm$ 13.43 & 1.08 $\pm$ 0.24 & 58.24 $\pm$ 9.10 & 2.09 $\pm$ 0.53 & 55.01 $\pm$ 11.49 & 2.54 & 45.67 \\
    
    &\textbf{\ShortName (Ours)}&\textbf{ 1.59$\pm$0.60} & \textbf{31.59$\pm$8.33} &  \textbf{0.77 $\pm$ 0.21} & \textbf{64.94 $\pm$ 11.73}  & \textbf{1.40 $\pm$ 0.41} & \textbf{67.98 $\pm$ 11.46} & \textbf{1.25} & \textbf{54.84} \\
    \hline
    
    \multirow{2}{*}{\textbf{4 views}}&\textbf{OccNet-C} & 5.51 $\pm$ 4.02 & 28.17 $\pm$ 13.88 & 1.03 $\pm$ 0.19 & 60.30 $\pm$ 6.62 & 1.95 $\pm$ 0.53 & 57.19 $\pm$ 12.48 & 2.83 & 48.55  \\
    
    &\textbf{\ShortName (Ours)} & \textbf{1.78$\pm$0.78} & \textbf{32.58$\pm$7.74} & \textbf{0.73$\pm$0.14} & \textbf{65.21 $\pm$ 5.93} & \textbf{1.34 $\pm$ 0.53} & \textbf{70.91 $\pm$ 12.32} & \textbf{1.28} & \textbf{56.23} \\
    \bottomrule \\
  \end{tabular}}
\caption{\small{\textbf{Articulation-canonicalized} shape reconstruction from different number of input views. Our \ShortName (Ours) consistently outperforms an OccNet\cite{OccNet}-based baseline in both CD and IoU over all categories.}}
\label{tab:canonical}
\end{table*}
We evaluate our method on the articulated objects test split. In Table~\ref{tab:canonical} we report the mesh reconstruction quality w.r.t the articulation-canonicalized mesh using 1, 2 and 4 input views. We use two standard metrics: 3D Intersection over Union (IoU), and Chamfer Distance (CD) between points sampled from the meshes, and note that CD is a better measure of the detail precision than IoU which instead provides a coarser estimate. This distinction is especially important when comparing shape in a canonical articulation where different instances of the same category have a significant overlap. It can be seen that \ShortName consistently outperforms the baseline OccNet-C on both metrics by a large margin across the different categories. Especially note the large improvement in CD. 


In Figure~\ref{fig:visualresults} we visualize the reconstruction quality of 2 instances from each category by OccNet-C and our \ShortName. As can be seen, \ShortName manages to capture detailed geometric properties of the input instance. In contrast, OccNet-C seems to neglect small-scale details as especially emphasized on the laptop keyboard and oven panel. In the glasses category, while OccNet-C  generates plausible outputs, we noticed that they often do not match the true object geometry. This further highlights the previously discussed difference between CD and IoU. A possible explanation for the ``regression to the mean'' effect seen on OccNet-C is that supervision is done only at a canonical pose.
\ShortName on the other hand benefits from its structured prediction that enforces attention to details for the explicit canonicalization to succeed.

\renewcommand{\arraystretch}{1.4}
\setlength{\tabcolsep}{5pt}
\begin{table*}[t!]
\centering
\centering
\footnotesize
\resizebox{0.9\linewidth}{!}{
\begin{tabular}{llcccccc|cc}
    \toprule
    &\textbf{Class} & \multicolumn{2}{c}{\textbf{eyeglasses}} &  \multicolumn{2}{c}{\textbf{laptop}} &  \multicolumn{2}{c}{\textbf{oven}}  &  \multicolumn{2}{c}{\textbf{mean}}\\    
    &\textbf{Metric} & \textbf{CD~$\downarrow$} & \textbf{IoU~$\uparrow$} & \textbf{CD~$\downarrow$} & \textbf{IoU~$\uparrow$} & \textbf{CD~$\downarrow$} & \textbf{IoU~$\uparrow$} &
    \textbf{CD~$\downarrow$} & \textbf{IoU~$\uparrow$}\\
    \midrule
    \multirow{2}{*}{\textbf{Single view}}&\textbf{OccNet-F}  & 9.65 $\pm$ 2.69 &  \textbf{19.97 $\pm$ 11.73} & 1.42 $\pm$ 0.27 & 52.31 $\pm$ 9.00 & 5.00 $\pm$ 1.93 & 51.72 $\pm$ 12.91 & 5.36 & 41.33 \\
    
    &\textbf{\ShortName} & \textbf{4.19 $\pm$ 2.13} & 13.54 $\pm$ 8.02 & \textbf{1.08 $\pm$ 0.31} & \textbf{56.17 $\pm$ 11.06} & \textbf{2.12 $\pm$ 1.42} & \textbf{61.32 $\pm$ 9.17} & \textbf{2.46} & \textbf{43.67} \\
    \hline
    
    \multirow{2}{*}{\textbf{2 views}}&\textbf{OccNet-F} & 8.49 $\pm$ 2.37 & \textbf{27.77 $\pm$ 8.59} & 1.75 $\pm$ 0.47 & 45.77 $\pm$ 9.58 & 4.82 $\pm$ 1.78 & 50.25 $\pm$ 10.36 & 5.02 & 41.26 \\
    
    &\textbf{\ShortName}& \textbf{2.16 $\pm$ 0.81} & 21.26 $\pm$ 8.83 & \textbf{1.23 $\pm$ 0.52} & \textbf{50.71 $\pm$ 11.96}  & \textbf{1.95 $\pm$ 0.89}& \textbf{60.86 $\pm$ 8.87}  & \textbf{1.78} & \textbf{44.28} \\
    \hline
    
    \multirow{2}{*}{\textbf{4 views}}&\textbf{OccNet-F} & 9.10 $\pm$ 2.69 & 28.20 $\pm$ 10.09 & 1.36 $\pm$ 0.30 & 56.98 $\pm$ 7.88 & 3.13 $\pm$ 1.21 & 52.95 $\pm$ 10.41 &  4.53 & 46.04 \\
    
    &\textbf{\ShortName} & \textbf{2.00 $\pm$ 0.74} & \textbf{31.13 $\pm$ 8.50} & \textbf{1.06 $\pm$ 0.34} & \textbf{57.85 $\pm$ 10.34} & \textbf{1.95 $\pm$ 0.67} & \textbf{63.31 $\pm$ 8.15} & \textbf{1.67} & \textbf{50.76} \\
    \bottomrule \\
  \end{tabular}}
\caption{\small{\textbf{Reconstruction in Input Shape} shape reconstruction from different number of input views. Our \ShortName consistently outperforms an OccNet\cite{OccNet}-based baseline in both CD and IoU.}}
\label{tab:input_pose}
\end{table*}
In Table~\ref{tab:input_pose} we evaluate \ShortName on reconstructing in the input pose. We emphasize that while existing methods (e.g. \cite{OccNet}) can reconstruct 3D shape from single images, they cannot aggregate multiple views to complete the geometry and instead hallucinate it. \ShortName, in contrast, is specially designed to do so. This is demonstrated by the markedly lower reconstruction errors compared with OccNet-F. Albeit strictly superior wrt the baseline, interestingly \ShortName seems to perform worse when measured on the input pose than on the canonical pose. We attribute this to the error in reposing. Note, however, that this error does not reflect a loss of details (see Figure \ref{fig:repose_results}) but rather that a small but consistent mismatch in pose (e.g. laptop lid angle) would cause  large reconstruction errors.

\subsection{Ablation study}\label{subsec:ablation}
\renewcommand{\arraystretch}{1}
\setlength{\tabcolsep}{5pt}
\begin{table*}[th!]
\centering
\centering
\footnotesize
\resizebox{0.9\linewidth}{!}{
\begin{tabular}{lcccccc|cc}
    \toprule
    \textbf{Class}     & \multicolumn{2}{c}{\textbf{eyeglasses}} &  \multicolumn{2}{c}{\textbf{laptop}} &  \multicolumn{2}{c}{\textbf{oven}}    & \multicolumn{2}{c}{\textbf{mean}}\\    

    \textbf{Metric}   & \textbf{CD~$\downarrow$} & \textbf{IoU~$\uparrow$} & \textbf{CD~$\downarrow$} & \textbf{IoU~$\uparrow$} & \textbf{CD~$\downarrow$} & \textbf{IoU~$\uparrow$} &
    \textbf{CD~$\downarrow$} & \textbf{IoU~$\uparrow$}\\
    \midrule
    \textbf{Ours} & \textbf{1.59$\pm$0.60 }  & \textbf{31.59$\pm$8.33} & \textbf{0.77 $\pm$ 0.21} & \textbf{64.94 $\pm$ 11.73} & 1.40 $\pm$ 0.41& \textbf{67.98 $\pm$ 11.46} & \textbf{1.25} & \textbf{54.84}  \\
    
    \textbf{w/o consistency} & 2.09 $\pm$ 0.81 &  29.54 $\pm$ 9.73 & 0.83 $\pm$ 0.20 &  60.03 $\pm$ 9.34 & 1.42 $\pm$ 0.36 &  63.83 $\pm$ 7.26 & 1.44 & 51.13  \\
    
    \textbf{w/o local feature} & 2.53 $\pm$ 1.71 &  25.94 $\pm$ 10.28 & 0.81 $\pm$ 0.22 &  61.66 $\pm$ 11.33 & \textbf{1.38 $\pm$ 0.49} & 65.81 $\pm$ 10.47 & 1.98 & 51.14  \\
    
    
    \bottomrule \\
  \end{tabular}

}
\caption{{\small Our ablation on two pipeline ingredient: consistency loss and local features  shows their effectiveness in improving overall reconstruction accuracy.}}
\label{tab:ablation}
\end{table*}
\parahead{Design choices}
In Section~\ref{sec:method} we discuss two main ingredients of our network architecture: (1)~consistency loss: used to regularize the canonical reconstructions from different views; and (2)~local features: used to propagate image cues through the lifted points. We evaluate their contribution to the overall performance of \ShortName in Table~\ref{tab:ablation} by comparing the reconstruction quality of our method in its complete form (top row) with its reduced variants. As can be seen, both ingredients play an important role in providing an overall improved performance. 

\parahead{Adding views at inference}
A property of our pipeline is that thanks to canonicalization, it can support an arbitrary number of input views at inference even if trained on a smaller set size. 
This is confirmed by Figure \ref{fig:multiview_inference_compact} where we ran inference on 1 to 6 input views using a \ShortName trained on only 2 views. As can be seen the reconstruction performance keeps increasing with the addition of input views.


\subsection{Reposing}\label{subsec:reposing}
\vspace{-0.1in}
\begin{figure}
\begin{center}
\includegraphics[width=\linewidth]{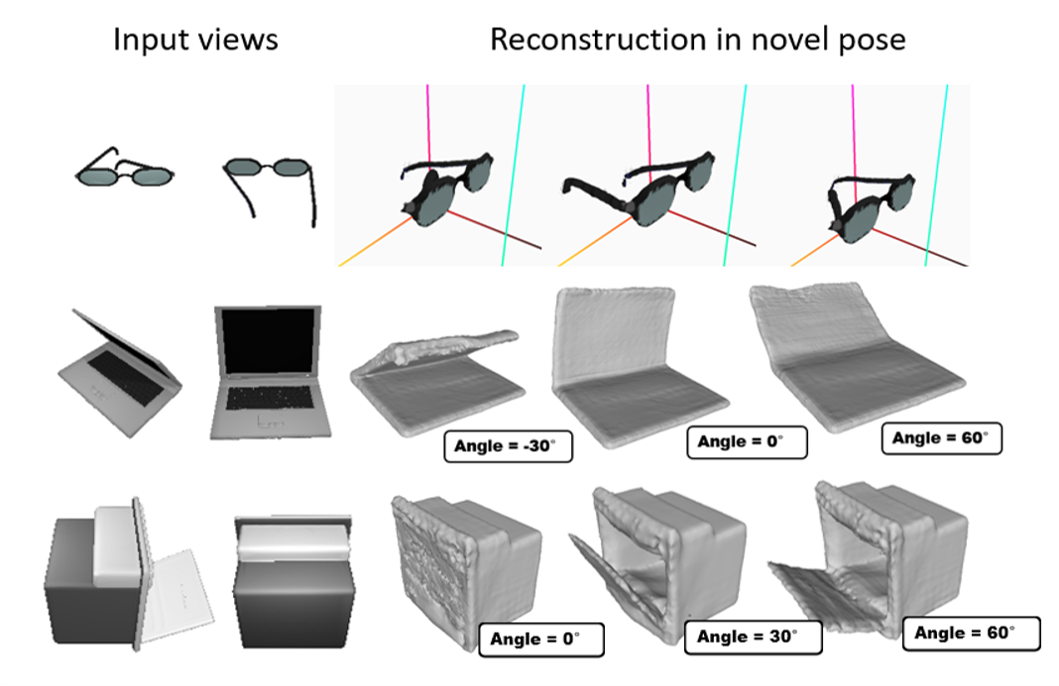}
\end{center}
   \caption{\textbf{Reposing.} \ShortName outputs animatable reconstructions demonstrated here on 3 different poses.}
\label{fig:repose_results}
\end{figure}

A unique feature of \ShortName is its animatable output. By keeping the intermediate representations of the reconstructed 3D shape as a union of lifted pointclouds and joint predictions, we allow user-provided pose input to repose the pointclouds before passed to the reconstruction module. Since the reconstruction module is trained to reconstruct at input poses, it can gracefully support this application. 
In Figure~\ref{fig:repose_results}, we visualize the reposing of a reconstruction from 2 input views, into multiple output articulations. We implement two types of reposing. The first, which we demonstrate on the laptop and oven categories (bottom two rows) follows the network reposing explained in section~\ref{sec:method}, namely: reposing the pointcloud from which a reconstruction is performed. The second reposing technique, which we demonstrate using the eyeglasses, reposes the mesh directly by starting at the canonical mesh reconstruction and modifying each vertex according to its nearest neighbor canonical point cloud joint assignment. We invite the reader to check the supplementary for more animations.

\section{Limitations \& Future Work}
\vspace{-0.1in}
Our approach has several limitations. First, due to limited training data, we focused on synthetic animatable datasets. Extending our work to real scenes and is an important step forward. In Figure \ref{fig:teaser} we show first encouraging results on real data. Second, we assume a fixed number of joints per category, but this is not always true (\eg,~chairs that can swivel, fold, etc.). Lastly, there is a large class on non-rigid objects (e.g. clothing) whose deformation cannot be efficiently modeled using articulations. We believe that multi-view aggregation of deformable objects from sparse input view is an important research direction and hope to see more research done along this line.

{\small
\bibliographystyle{ieee_fullname}
\bibliography{references}
}

\end{document}